# Context-Guided Dynamic Retrieval for Improving Generation Quality in RAG Models

Jacky He
Cornell University
New York, USA

Guiran Liu
San Francisco State University
San Francisco, USA

Binrong Zhu
San Francisco State University
San Francisco, USA

Hanlu Zhang
Stevens Institute of Technology
Hoboken, USA

Hongye Zheng
The Chinese University of Hong Kong
Hong Kong, China

Xiaokai Wang*
Santa Clara University
Santa Clara, USA

*Abstract*-This paper focuses on the dynamic optimization of the Retrieval-Augmented Generation (RAG) architecture. It proposes a state-aware dynamic knowledge retrieval mechanism to enhance semantic understanding and knowledge scheduling efficiency in large language models for open-domain question answering and complex generation tasks. The method introduces a multi-level perceptive retrieval vector construction strategy and a differentiable document matching path. These components enable end-to-end joint training and collaborative optimization of the retrieval and generation modules. This effectively addresses the limitations of static RAG structures in context adaptation and knowledge access. Experiments are conducted on the Natural Questions dataset. The proposed structure is thoroughly evaluated across different large models, including GPT-4, GPT-4o, and DeepSeek. Comparative and ablation experiments from multiple perspectives confirm the significant improvements in BLEU and ROUGE-L scores. The approach also demonstrates stronger robustness and generation consistency in tasks involving semantic ambiguity and multi-document fusion. These results highlight its broad application potential and practical value in building high-quality language generation systems.

## I. INTRODUCTION

With the widespread application of Large Language Models (LLMs) in natural language processing, they have demonstrated outstanding performance in tasks such as text generation, question answering, and dialogue understanding. However, a key issue with these models lies in their closed knowledge. Since the model parameters are fixed after training, updating knowledge requires retraining or fine-tuning. This becomes particularly inadequate in scenarios where information changes rapidly or domain knowledge evolves continuously [1]. To overcome this limitation, researchers proposed the Retrieval-Augmented Generation (RAG) framework. This framework integrates a document retrieval module with a generation model, allowing the model to dynamically access external knowledge bases during generation. This integration significantly improves factual accuracy and timeliness, making it a crucial direction for enhancing language model capabilities [2].

Traditional RAG structures mainly rely on static retrieval mechanisms. After receiving an input query, they retrieve relevant documents from a pre-built vector index and feed them into the generation model as context. Although this approach partially alleviates the issue of outdated knowledge, its fixed retrieval strategy struggles to meet complex and evolving semantic needs. This limitation affects both the reasoning capacity and expression precision of the generation model. In particular, tasks such as complex question answering, multi-turn dialogues, or cross-domain information integration often require high-quality and dynamic knowledge support, which static retrieval fails to provide. Therefore, building a more flexible, efficient, and adaptive retrieval mechanism has become key to further developing RAG structures [3].

The introduction of dynamic knowledge retrieval brings new opportunities for improving RAG structures. Unlike traditional methods, dynamic mechanisms emphasize updating the retrieval strategy continuously during generation [4-6]. They adjust retrieval goals and semantic vector spaces in real time, based on the current generation state, contextual information, and knowledge gaps. This approach improves the accuracy of knowledge utilization and enhances the model's interactivity and semantic understanding. In cutting-edge applications such as multimodal generation [7], domain-specific generation [8], recommendation systems [9], and zero-shot learning [10], dynamic retrieval has the potential to greatly expand the applicability and quality of language models [11]. Thus, exploring how to effectively integrate dynamic retrieval mechanisms into the RAG framework is both theoretically meaningful and practically valuable.

At the same time, current mainstream RAG systems still suffer from a degree of disconnection between retrieval and generation. Although the retrieval module can provide highly relevant external documents, the generation module may fail to fully leverage the retrieved content. This results in redundancy, ambiguity, or even distortion in the generated output. To address this issue, an efficient mechanism for knowledge selection, fusion, and dynamic feedback is urgently needed[12]. Such a mechanism would enable true collaborative optimization in the retrieval-generation loop. In addition, existing RAG models are limited by static input representations. They lack the ability to flexibly perceive user intent and adapt to shifting knowledge demands. Therefore, improving RAG with dynamic knowledge retrieval is expected to play a key role in enhancing semantic understanding, contextual reasoning, and generation accuracy [13].

In summary, the RAG framework has laid the foundation for enhancing the open knowledge acquisition capabilities of language models. Introducing dynamic knowledge retrieval further drives the framework toward greater efficiency, intelligence, and personalization. As AI systems move toward large-scale applications, complex interactions, and knowledge-intensive tasks [14-16], dynamically improving the RAG architecture holds significant theoretical and practical importance. This study aims to explore and implement a dynamic RAG approach tailored for complex generation tasks. It focuses on solving core issues in knowledge utilization efficiency, contextual adaptability, and quality control, providing both theoretical grounding and technical support for building more intelligent and general-purpose language generation systems.

## II. METHOD

This study introduces a dynamic knowledge retrieval mechanism based on the classic RAG framework to achieve real-time call and adaptive scheduling of external knowledge during the generation process. The model architecture is shown in Figure 1.

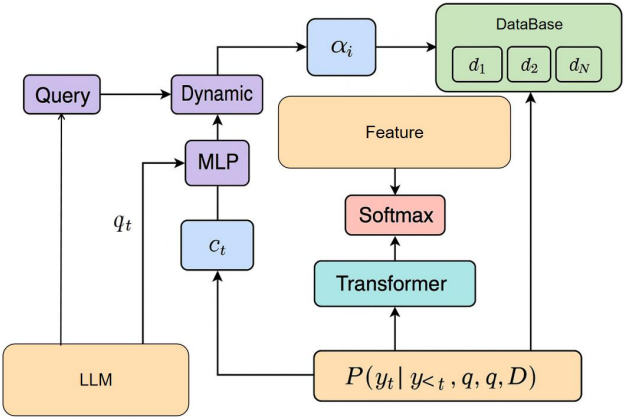

Figure 1. Overall model architecture

The model architecture diagram illustrates an enhanced Retrieval-Augmented Generation (RAG) structure incorporating a dynamic knowledge retrieval mechanism. Inspired by the hierarchical semantic encoding strategies proposed by Cai et al. [17], the model employs a multi-level perceptive retrieval vector construction, enabling fine-grained document understanding and context-sensitive retrieval. To ensure tight coupling between retrieval and generation, we adopt a differentiable document matching path that allows gradient flow across modules, drawing on techniques validated by Yu et al. [18] in joint retrieval and generation frameworks for improved semantic consistency. This differentiability supports end-to-end joint training, encouraging retrieval vectors to evolve alongside generation objectives. Additionally, the dynamic scheduling of retrieval steps is informed by the model compression and tuning techniques outlined by Kai, Zhu, and Gong [19], which help balance retrieval quality and latency under large-scale deployment scenarios. These elements collectively form the basis of our improved RAG structure. The overall process starts with the query vector q, generates a state-aware retrieval vector $q_t^{'}$ through MLP and dynamic modules, and retrieves the most relevant document vector set from the external database. Subsequently, the retrieval score $a_i$ guides the document fusion to form a context embedding $c_i$, which is jointly input into the Transformer with the LLM generation history to output the probability distribution of the current generated word. This structure achieves dynamic coupling between retrieval and generation through the differentiable retrieval path, attention control mechanism and generator collaborative optimization, effectively improving the efficiency of knowledge retrieval and generation quality.

We set the input as the query vector $q \in R^d$. The goal is to dynamically select the relevant document set $D_q \subset D$ from the knowledge base $D = \{d_1, d_2, ..., d_N\}$ through the retrieval module and fuse it into the generation model for text generation. First, we design a state-aware retrieval controller to adjust the retrieval vector $q_t^{'}$ according to the current context state $h_t$. The vector is obtained by jointly modeling the original query vector and the generated context:

$$q_t^{'} = MLP([q; h_t])$$

Among them, $[q; h_t]$ represents vector concatenation, MLP is a multi-layer perceptron module, and the output is an enhanced semantic vector, which is used to further improve the dynamic perception ability of knowledge fragments.

In the retrieval stage, we use a differentiable similarity estimation method for document matching [20], and calculate the relevance score between vector $q_t^{'}$ and document representation $d_i$ by scaled dot-product attention:

$$a_i = \frac{\exp(q_t^{'} \cdot d_i / \sqrt{d})}{\sum_{j=1}^{N} \exp(q_t^{'} \cdot d_j / \sqrt{d})}$$

Where $a_i$ represents the retrieval probability of the i-th document, $d_i \in R^d$ is the pre-encoded document vector, and the context document embedding $c_t = \sum_i a_i d_i$ is obtained by vector weighting to guide subsequent generation.

The generation module adopts a conditional language modeling strategy, which combines the currently generated hidden state $h_t$ and the retrieved knowledge context $c_t$ at each generation step to jointly predict the probability distribution of the next word. The generator is implemented through the Transformer structure, and the core prediction process is as follows:

$$P(y_t | y_{<t}, q, D) = \text{softmax}(W_o \cdot \text{Transformer}(y_{<t}, c_t))$$

Where $y_{<t}$ is the previously generated word sequence and $W_o$ is the output mapping matrix. By introducing the dynamic retrieval representation $c_t$, the model has the ability to dynamically adjust the knowledge input according to the generation history, avoiding the context drift caused by static document input.

In order to achieve end-to-end optimization of the retrieval module and the generation module, we designed a joint training loss function, which consists of a generation loss $L_{gen}$ and a retrieval contrast loss $L_{ret}$. The generation loss uses the standard cross entropy form, while the retrieval loss uses a contrastive learning framework to model positive and negative document pairs. The joint objective function is as follows:

$$L_{total} = L_{gen} + \lambda \cdot L_{ret}$$

The hyperparameter $\lambda$ controls the weight balance of the two losses. Through joint optimization, the model can not only learn stronger text generation capabilities, but also improve the ability to select knowledge under different semantic states, achieving efficient collaboration between retrieval and generation.

### III. EXPERIMENT

#### A. Datasets

This study selects Natural Questions (NQ) as the primary experimental dataset. Released by Google, NQ is designed for open-domain question answering tasks. It contains a large number of real user queries from search engines, each paired with corresponding Wikipedia documents and both short and long answers. The structure of NQ closely reflects real-world scenarios of retrieval-augmented generation. It effectively tests a model's ability in complex semantic understanding and knowledge retrieval.

In this work, we treat the questions in NQ as input queries q, and the associated Wikipedia passages as the external knowledge base D. The provided gold passage is used as the supervision signal. We use a DPR-based retriever to initialize the vector index for document pre-encoding. This ensures semantic consistency in document representations. Our dynamic retrieval mechanism is then applied to update vectors and model their matching process. To better support long-form generation and multi-passage fusion, we segment the original texts and unify the encoding format.

The dataset encompasses a wide range of query intents, extensive document spans, and substantial noise. These characteristics pose significant challenges in evaluating retrieval-augmented generation systems in realistic scenarios. To ensure reproducibility and comparability of results, we strictly adhere to the official train, validation, and test splits provided by NQ. Additionally, this verification process validates the effectiveness and generalizability of our proposed dynamic RAG structure in open-domain QA tasks.

#### B. Experimental Results

To assess the performance of the dynamic retrieval-augmented generation (RAG) structure across various large language models (LLMs), we conduct a model capability comparison experiment. We maintain the same retrieval mechanism and input parameters while introducing several prominent LLMs as the generation module. These include GPT-3.5, GPT-4, Qwenmax, DeepSeek, and the most recent GPT-4o. By comparing the generation quality, answer accuracy, and knowledge integration on the Natural Questions dataset, we analyze each model's response capability and generation robustness during dynamic knowledge access. This reveals how different models behave under the same retrieval-enhanced conditions. In particular, GPT-4o, as a new-generation multimodal model, shows strong performance in handling complex queries and long-context structures. It provides higher accuracy and better consistency. This offers solid support for applying dynamic RAG structures in real-world scenarios, and the actual results are shown in Table 1.

Table 1. Comparative test results of different large models

| Model | BLEU | ROUGE-L | Average response time (ms) |
|---|---|---|---|
| GPT-3.5[21] | 34.2 | 47.5 | 920 |
| Qwenmax[22] | 38.9 | 52.3 | 980 |
| Deepseek[23] | 39.1 | 53.5 | 890 |
| GPT4[21] | 40.2 | 54.3 | 1010 |
| GPT4o[21] | 41.7 | 55.1 | 990 |

The experimental results show that all large language models demonstrate relatively stable generation performance under the dynamic RAG structure. Both BLEU and ROUGE-L scores tend to increase with model capability. GPT-3.5, as an earlier version, shows weaker performance, with a BLEU score of 34.2 and a ROUGE-L score of 47.5, significantly lower than the later models. In contrast, Qwenmax and DeepSeek, two mid-sized emerging models, perform well in generation quality. Notably, DeepSeek achieves a ROUGE-L score of 53.5, indicating strong abilities in semantic preservation and contextual continuity.

GPT-4 and GPT-4o, as more advanced models, show further improvements in overall performance. GPT-4o achieves the best results on both BLEU (41.7) and ROUGE-L (55.1), highlighting its strength in complex query understanding and language generation. It is worth noting that while GPT-4 scores slightly lower than GPT-4o, its average response time reaches

1010 ms, the highest among all models. This suggests a larger inference latency, possibly due to model complexity or resource scheduling bottlenecks.

Overall, GPT-4o strikes a good balance between generation quality and response efficiency. It produces more accurate and consistent outputs, while also outperforming GPT-4 and Qwenmax in average response time. This indicates that under dynamic knowledge retrieval, both reasoning ability and real-time responsiveness jointly affect the final generation quality. Therefore, in practical deployment, model performance, resource consumption, and interaction speed should all be considered to select the most suitable generation engine for the target application scenario.

At the same time, this paper also gives the experimental results of analyzing the impact of different retrieval vector construction methods on the generation quality, and the experimental results are shown in Figure 2.

The method proposed in this study (Ours) builds upon these insights by introducing an attention-based fusion mechanism to construct retrieval vectors. This design allows the model to effectively integrate both query semantics and contextual signals when retrieving external documents. As a result, it achieves the highest performance across both evaluation metrics, with a BLEU score of 41.0 and a ROUGE-L score of 53.0. These results validate the proposed method's strength in modeling contextual relevance and dynamically scheduling knowledge throughout the generation process. By enabling more precise control over what information is retrieved and how it is used, this approach significantly enhances the accuracy, fluency, and overall quality of the generated text.

Finally, this paper presents the experimental results of dynamic RAG robustness evaluation on a data subset based on query ambiguity, and the experimental results are shown in Figure 3.

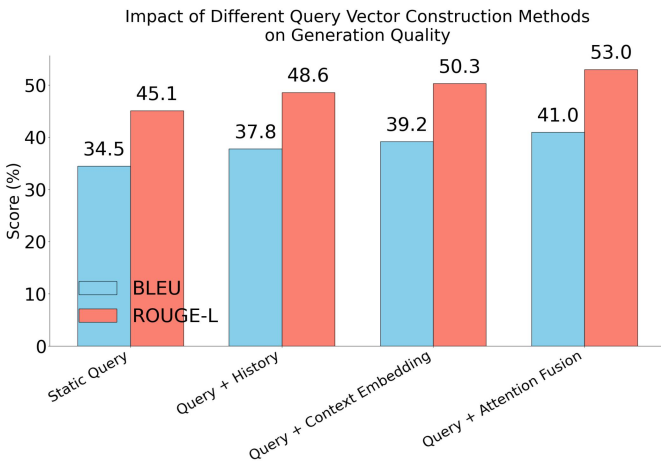

Figure 2. Histogram of the experiment "Analysis of the impact of different retrieval vector construction methods on generation quality"

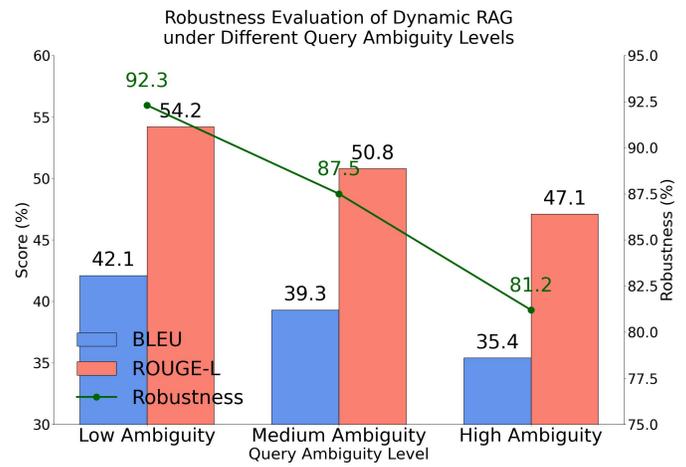

Figure 3. Robustness Evaluation of Dynamic RAG under Different Query Ambiguity Levels

The experimental results in the figure show a consistent improvement in model performance on BLEU and ROUGE-L scores as the construction of retrieval vectors becomes more advanced. The basic static query method achieves the lowest scores, with 34.5 on BLEU and 45.1 on ROUGE-L. This indicates clear limitations in semantic expression and contextual consistency, making it difficult to meet the needs of dynamic generation.

When historical information and contextual embeddings are incorporated, the model's ability to understand the semantics of the query is noticeably enhanced. This improvement is reflected in the upward trend of BLEU and ROUGE-L scores. The results indicate that richer semantic representations provide stronger support for generating more coherent and contextually appropriate outputs. Specifically, the Query + Context Embedding approach demonstrates a distinct advantage in maintaining semantic continuity across the generated content. This is particularly evident in the ROUGE-L score, which shows a clear improvement compared to other settings.

The experimental results show that as query ambiguity increases, the dynamic RAG model exhibits a noticeable decline in both generation quality and robustness. In low-ambiguity scenarios, the model can fully understand the query intent. BLEU and ROUGE-L scores reach 42.1 and 54.2, respectively, with robustness at 92.3%. This indicates that the generation system can reliably access external knowledge and produce accurate outputs under clear query conditions.

When ambiguity rises to a moderate level, model performance starts to decline. BLEU drops to 39.3, ROUGE-L falls to 50.8, and robustness decreases to 87.5%. This suggests that under less explicit intent or semantic ambiguity, the retrieval module struggles to select precise knowledge segments. As a result, the stability and coherence of the generated text are compromised.

In high-ambiguity scenarios, the scores decline further. BLEU drops to 35.4, and robustness falls to 81.2%. These results show that the current dynamic retrieval mechanism remains sensitive to semantic uncertainty. Although it offers some adaptive capacity, it still lacks strong context awareness and effective disambiguation strategies. Enhancing these aspects is essential for improving robustness and generation consistency when dealing with vague or ambiguous queries.

IV. CONCLUSION

This study focuses on the dynamic retrieval mechanism within the Retrieval-Augmented Generation (RAG) architecture. The goal is to improve the efficiency of knowledge utilization and semantic adaptability in large language models during generation tasks. By introducing a state-aware retrieval vector construction method and a differentiable document matching pathway, the retrieval and generation processes are jointly optimized. Experiments conducted on the Natural Questions dataset provide a comprehensive evaluation, confirming the superiority of the proposed dynamic RAG structure in handling complex semantic queries, multi-document fusion, and generation consistency. In particular, comparative experiments with various large language models demonstrate that incorporating a dynamic retrieval strategy significantly boosts key metrics such as BLEU and ROUGE-L, enhancing the model's robustness and output quality in real-world scenarios.

Additionally, ablation studies on the retrieval vector construction and robustness assessments under ambiguous queries further highlight the generality and adaptability of the proposed method. Under complex contexts and vague intentions, the dynamic RAG structure maintains high response accuracy and content coherence. This indicates strong capabilities in semantic comprehension and dynamic coordination. In conclusion, this work not only deepens the theoretical understanding of dynamic retrieval mechanisms within the RAG framework but also provides a solid technical foundation and experimental evidence for building more intelligent and real-time generative systems.

REFERENCES

[1] Z. Hei, et al., "Dr-rag: Applying dynamic document relevance to retrieval-augmented generation for question-answering," arXiv preprint arXiv:2406.07348, 2024.

[2] W. Su, et al., "DRAGIN: Dynamic Retrieval Augmented Generation based on the Information Needs of Large Language Models," arXiv preprint arXiv:2403.10081, 2024.

[3] Y. Wu, Y. Fang and L. Liao, "Retrieval Augmented Generation for Dynamic Graph Modeling," arXiv preprint arXiv:2408.14523, 2024.

[4] W. Huang, J. Zhan, Y. Sun, X. Han, T. An and N. Jiang, "Context-Aware Adaptive Sampling for Intelligent Data Acquisition Systems Using DQN," arXiv preprint arXiv:2504.09344, 2025.

[5] F. Guo, X. Wu, L. Zhang, H. Liu and A. Kai, "A Self-Supervised Vision Transformer Approach for Dermatological Image Analysis," Journal of Computer Science and Software Applications, vol. 5, no. 4, 2025.

[6] Y. Deng, "A Reinforcement Learning Approach to Traffic Scheduling in Complex Data Center Topologies," Journal of Computer Technology and Software, vol. 4, no. 3, 2025.

[7] J. Zhan, "Single-Device Human Activity Recognition Based on Spatiotemporal Feature Learning Networks," Transactions on Computational and Scientific Methods, vol. 5, no. 3, 2025.

[8] S. Wang, Z. Liu and B. Peng, "A Self-training Framework for Automated Medical Report Generation," Proceedings of the 2023 Conference on Empirical Methods in Natural Language Processing, pp. 16443-16449, December 2023.

[9] L. Zhu, "Deep Learning for Cross-Domain Recommendation with Spatial-Channel Attention," Journal of Computer Science and Software Applications, vol. 5, no. 4, 2025.

[10] Z. Liu, M. Wu, B. Peng, Y. Liu, Q. Peng and C. Zou, "Calibration Learning for Few-shot Novel Product Description," Proceedings of the 46th International ACM SIGIR Conference on Research and Development in Information Retrieval, pp. 1864-1868, July 2023.

[11] J. Kim, D. Ko and G. Kim, "DynamicER: Resolving Emerging Mentions to Dynamic Entities for RAG," arXiv preprint arXiv:2410.11494, 2024.

[12] X. Yan, J. Du, L. Wang, Y. Liang, J. Hu and B. Wang, "The Synergistic Role of Deep Learning and Neural Architecture Search in Advancing Artificial Intelligence", Proceedings of the 2024 International Conference on Electronics and Devices, Computational Science (ICEDCS), pp. 452-456, Sep. 2024.

[13] Y. Shi, et al., "Enhancing retrieval and managing retrieval: A four-module synergy for improved quality and efficiency in rag systems," arXiv preprint arXiv:2407.10670, 2024.

[14] X. Li, Y. Peng, X. Sun, Y. Duan, Z. Fang and T. Tang, "Unsupervised Detection of Fraudulent Transactions in E-commerce Using Contrastive Learning," arXiv preprint arXiv:2503.18841, 2025.

[15] Y. Zhang, "Social Network User Profiling for Anomaly Detection Based on Graph Neural Networks," arXiv preprint arXiv:2503.19380, 2025.

[16] A. Liang, "A Graph Attention-Based Recommendation Framework for Sparse User-Item Interactions," Journal of Computer Science and Software Applications, vol. 5, no. 4, 2025.

[17] G. Cai, J. Gong, J. Du, H. Liu and A. Kai, "Investigating Hierarchical Term Relationships in Large Language Models," Journal of Computer Science and Software Applications, vol. 5, no. 4, 2025.

[18] Z. Yu, S. Wang, N. Jiang, W. Huang, X. Han and J. Du, "Improving Harmful Text Detection with Joint Retrieval and External Knowledge," arXiv preprint arXiv:2504.02310, 2025.

[19] A. Kai, L. Zhu and J. Gong, "Efficient Compression of Large Language Models with Distillation and Fine-Tuning," Journal of Computer Science and Software Applications, vol. 3, no. 4, pp. 30–38, 2023.

[20] J. Wei, Y. Liu, X. Huang, X. Zhang, W. Liu and X. Yan, "Self-Supervised Graph Neural Networks for Enhanced Feature Extraction in Heterogeneous Information Networks", 2024 5th International Conference on Machine Learning and Computer Application (ICMLCA), pp. 272-276, 2024.

[21] P. A. Massey, C. Montgomery and A. S. Zhang, "Comparison of ChatGPT–3.5, ChatGPT-4, and orthopaedic resident performance on orthopaedic assessment examinations," JAAOS - Journal of the American Academy of Orthopaedic Surgeons, vol. 31, no. 23, pp. 1173–1179, 2023.

[22] L. Zhang, et al., "Qwen-IG: A Qwen-based Instruction Generation Model for LLM Fine-tuning," Proceedings of the 2024 13th International Conference on Computing and Pattern Recognition, 2024.

[23] A. Liu, et al., "Deepseek-v3 technical report," arXiv preprint arXiv:2412.19437, 2024.